%% file: root.tex
\documentclass[letterpaper, 10 pt, conference]{ieeeconf}  % Comment this line out if you need a4paper

\IEEEoverridecommandlockouts                              % This command is only needed if 
\usepackage{amsmath,amsfonts,amssymb}
\usepackage[dvipsnames]{xcolor}
\usepackage{tikz}
\usepackage{booktabs}
\usepackage{array}
\usepackage{algorithm}
\usepackage{algpseudocode}
\usetikzlibrary{calc, positioning, shapes}
\usepackage{graphicx}
\graphicspath{ {./gfx/} }

\usepackage{xcolor}

\title{\LARGE \bf
 Leveraging Multi-modal Sensing for Robotic Insertion Tasks in R\&D Laboratories
}

\author{Aaron Butterworth$^{1}$, Gabriella Pizzuto$^{2}$, Leszek Pecyna$^{3}$, Andrew I. Cooper$^{2}$ and Shan Luo$^{3}$ % <-this % stops a space
\thanks{$^{1}$A. Butterworth is with the Department of Computer Science, University of Liverpool, Liverpool L69 3BX, United Kingdom. E-mail: {\tt\small a.butterworth@liverpool.ac.uk}.}%
\thanks{$^{2}$G. Pizzuto and A.I. Cooper are with the Department of Chemistry, University of Liverpool, Liverpool L69 3BX, United Kingdom. E-mail: {\tt\small name.surname@liverpool.ac.uk}.}%
\thanks{$^{3}$S. Luo and L. Pecyna are with the Department of Engineering, King's College London, London WC2R 2LS, United Kingdom. E-mail: {\tt\small name.surname@kcl.ac.uk}.}%
}

\begin{document}

\maketitle
\thispagestyle{empty}
\pagestyle{empty}

%%%%%%%%%%%%%%%%%%%%%%%%%%%%%%%%%%%%%%%%%%%%%%%%%%%%%%%%%%%%%%%%%%%%%%%%%%%%%%%%
\begin{abstract}
Performing a large volume of experiments in Chemistry labs creates repetitive actions costing researchers time, automating these routines is highly desirable. Previous experiments in robotic chemistry have performed high numbers of experiments autonomously, however, these processes rely on automated machines in all stages from solid or liquid addition to analysis of the final product. In these systems every transition between machine requires the robotic chemist to pick and place glass vials, however, this is currently performed using open loop methods which require all equipment being used by the robot to be in well defined known locations. We seek to begin closing the loop in this vial handling process in a way which also fosters human-robot collaboration in the chemistry lab environment. To do this the robot must be able to detect valid placement positions for the vials it is collecting, and reliably insert them into the detected locations. We create a single modality visual method for estimating placement locations to provide a baseline before introducing two additional methods of feedback (force and tactile feedback). Our visual method uses a combination of classic computer vision methods and a CNN discriminator to detect possible insertion points, then a vial is grasped and positioned above an insertion point and the multi-modal methods guide the final insertion movements using an efficient search pattern. Through our experiments we show the baseline insertion rate of 48.78\% improves to 89.55\% with the addition of our `force and vision' multi-modal feedback method.
\end{abstract}

%%%%%%%%%%%%%%%%%%%%%%%%%%%%%%%%%%%%%%%%%%%%%%%%%%%%%%%%%%%%%%%%%%%%%%%%%%%%%%%%
\section{Introduction}
\input{new/intro}

\section{Related Works}
\input{new/background}

\section{Method}
\input{new/method}

\section{Experiments}
\input{new/experiments}

\section{Conclusion}
\input{new/conclusions}

%\footnotetext{The visual method cannot make additional placement attempts as it cannot detect a failed placement attempt before releasing the vial}
%\addtolength{\textheight}{-12cm}   % This command serves to balance the column lengths
                                  % on the last page of the document manually. It shortens
                                  % the textheight of the last page by a suitable amount.
                                  % This command does not take effect until the next page
                                  % so it should come on the page before the last. Make
                                  % sure that you do not shorten the textheight too much.

%%%%%%%%%%%%%%%%%%%%%%%%%%%%%%%%%%%%%%%%%%%%%%%%%%%%%%%%%%%%%%%%%%%%%%%%%%%%%%%%

%%%%%%%%%%%%%%%%%%%%%%%%%%%%%%%%%%%%%%%%%%%%%%%%%%%%%%%%%%%%%%%%%%%%%%%%%%%%%%%%

%%%%%%%%%%%%%%%%%%%%%%%%%%%%%%%%%%%%%%%%%%%%%%%%%%%%%%%%%%%%%%%%%%%%%%%%%%%%%%%%

%%%%%%%%%%%%%%%%%%%%%%%%%%%%%%%%%%%%%%%%%%%%%%%%%%%%%%%%%%%%%%%%%%%%%%%%%%%%%%%%

\bibliographystyle{IEEEtran}
\bibliography{bib.bib}
\end{document}

%% file: new/intro.tex
Automating a laboratory designed for humans poses several interesting challenges for robotics, a wide variety of transparent objects are employed and must be detected and handled in a safety-conscious manner. Creating collaborative environments for robotics and human scientists is an important problem, these collaborative lab spaces will significantly increase the speed at which experiments can be performed. A major challenge for these spaces is allowing robots to safely interact with glass labware, poor interactions can cause numerous hazards to the human collaborators and will cause disruption to the experiments in progress.
Therefore, equipping a laboratory robot with the ability to both see and touch when placing vials is a highly desirable requirement.

Although there has been an increase in usage of laboratory robotics~\cite{burger_mobile_2020, fakhruldeen2022, Shiri2021}, such workflows are still carried out in open loop. This hinders overall robot deployment for carrying out long-term laboratory experiments, such systems are vulnerable to disruption caused by a prior failure and pose a significant safety threat to human scientists should glassware break. Therefore, a robot in a chemistry lab, much like its human collaborators, should aim to use all possible sensory information to ensure a safe workflow and increased reliability. 
There exist different manual and tedious tasks across laboratory workflows that would greatly benefit from being automated; however, all of these would require manipulation of glass vials.

\begin{figure}[t!]
    \centering
    \includegraphics[width=.7\linewidth]{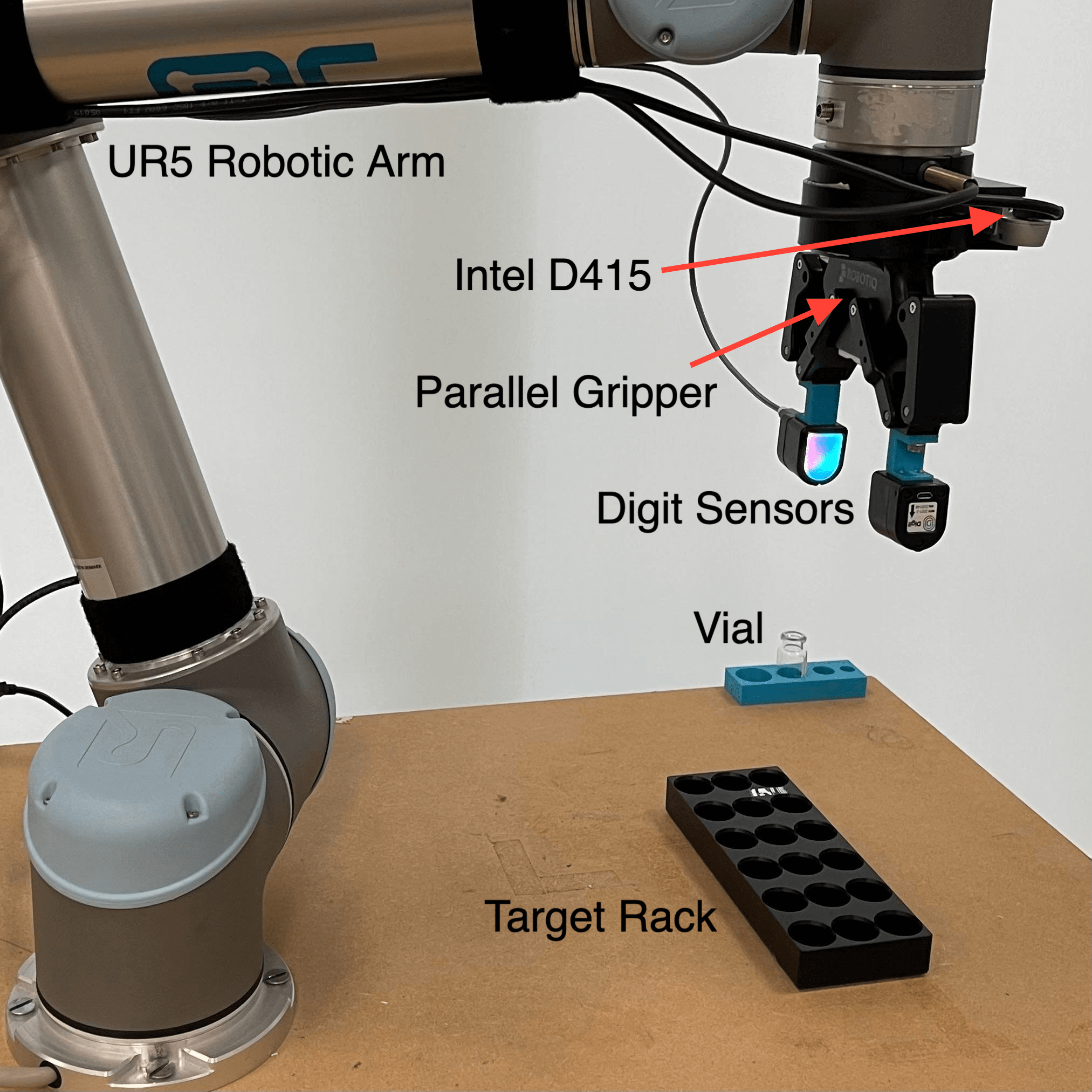}
    \caption{In the above laboratory automation environment, a robot arm equipped with a camera at its wrist and two camera-based tactile sensors~\cite{lambeta_digit_2020} attached to its 2-finger gripper is used to pick up a vial from a rack and insert it into a target rack.}
    \label{fig:setup}
\end{figure}
\begin{figure*}
    \centering
    \vspace{5pt}
    \includegraphics[trim=0 150 0 0, clip, width=.9\linewidth]{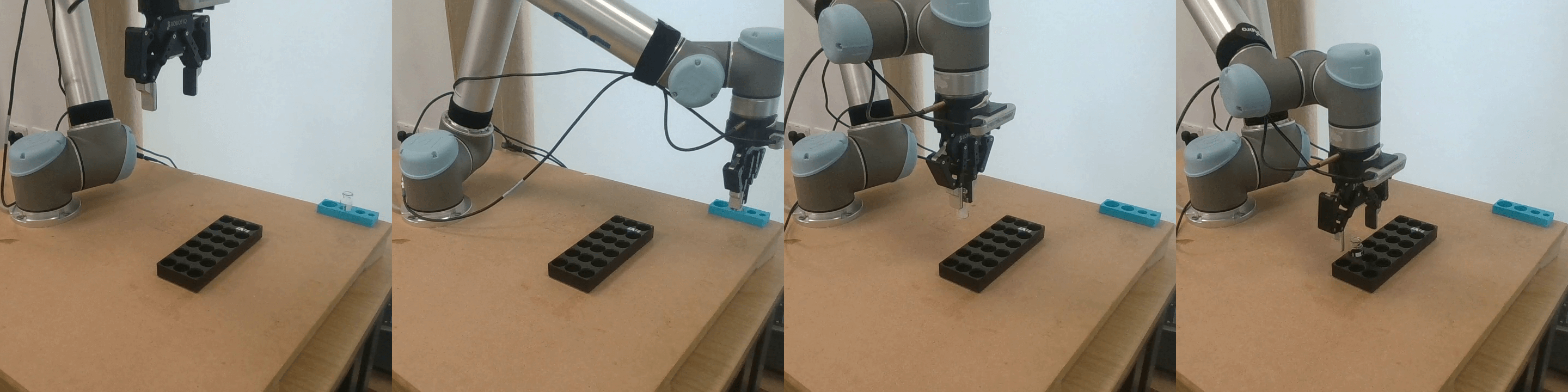}
    \caption{Sequential actions of the robot to perform vial insertion (from left to right): a). An image is taken of the workspace, the rack is detected and a target is chosen; b) The vial is collected; c). The vial is moved directly above the target; d). The vial is inserted.}
    \label{fig:walkthrough}
\end{figure*}

Up to date, tactile sensing has not been employed in laboratory automation. Existing robotic tasks rely on pre-programmed motions such as the mobile robotic chemist~\cite{burger_mobile_2020}. This rigidity presents an unnatural environment for a human collaborator, however, it provides high reliability which is highly desired for long experiments with many interactions. Multi-modal sensing has been introduced to object manipulation tasks outside of the laboratory setting to create a more sensor-rich environment and improve the reliability of object handling~\cite{pecyna_visual-tactile_2022, lee_making_2020}. By introducing multi-modal sensing we aim to allow the robot to operate in a dynamic environment more suitable for human-robot collaboration, while maintaining a high level of reliability suited to longer experiments.

In this paper, we introduce a novel multi-modal sensing method for robotic vial insertion, a common task in research laboratories. Our aim is to investigate the role of multi-modal sensory feedback in this task and how it can improve the performance of the robot from the visual baseline. In this task, we can represent almost any interaction between the robot and the labware as vials are loaded into and out of racks for transportation, from machines to racks, and from racks to machines.

As shown in Fig.~\ref{fig:setup}, in the simulated laboratory automation environment, we leverage the camera at the robot's wrist, intrinsic force sensors of the end-effector, two camera-based tactile sensors~\cite{lambeta_digit_2020} mounted onto its 2-finger gripper to improve the robotic insertion task. The experimental results show that our multi-modal approach boosts the success rate of the vial insertion to 89.55\%, compared to only 48.78\% with a single visual modality. The results show that multi-modal sensing can provide more cues of the interaction with the vial and the racks, and has potential to improve the reliability of a robotic system in the laboratory automation environment.

%% file: new/background.tex
\begin{figure*}
    \centering
    \vspace{5pt}
    \includegraphics[width=.6\linewidth]{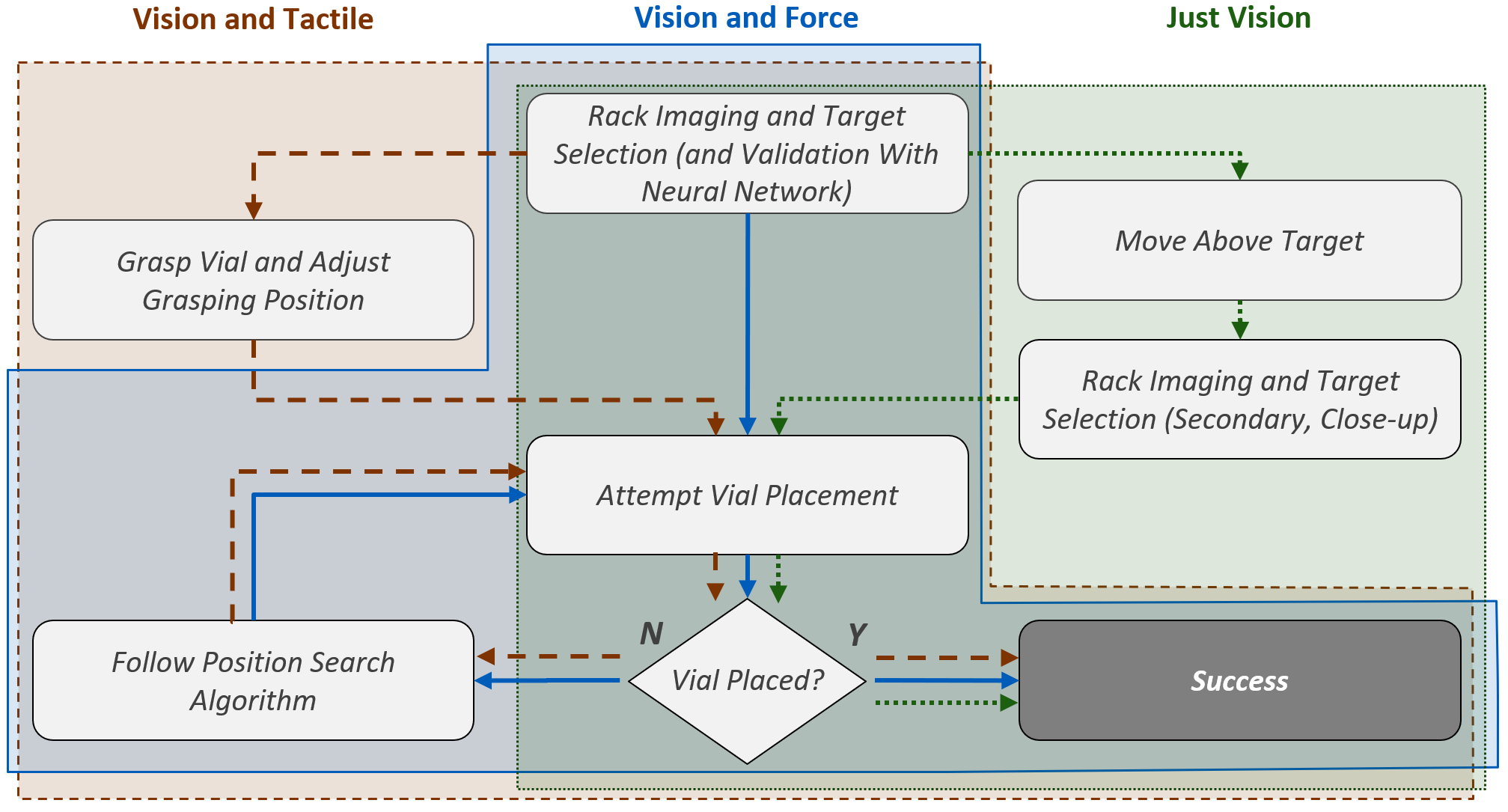}
    \caption{Workflow showing each of the three modalities. Note the method used to assess `Vial placed?' is different per system, however, its overall function in the workflow is the same.}
    \label{fig:flowchart}
\end{figure*}

\subsection{Laboratory Automation}
To better suit a changing laboratory workflow more versatile robotic systems are required.
Laboratory automation for material discovery has already taken advantage of different robotic platforms to carry out different workflows predominantly in the areas of materials for clean energy and pharmaceuticals.
Burger et al.~\cite{burger_mobile_2020} demonstrated that a mobile manipulator exceeds human-level performance for photocatalysis, where the robot carried out 688 experiments over seven days.  
The mobile robotic chemist freely moved around the lab space and requires much less modification to the existing human-oriented lab. 
However, this system still requires many assumptions as it operates in open loop, which reduces its generality. For example, the sample vials have to be of a set size and held in a specific rack which is then moved between fixed holders in the workspace; and the system is not capable of detecting these vials. 
As a result, a failed interaction may go unnoticed by the robot itself and thus jeopardises the remaining operations, while posing a safety hazard to human scientists.

Robotic manipulators have also been used for autonomously carrying out workflows to accelerate material discovery, for example, ~\cite{Shiri2021} and ~\cite{Pizzuto2022}.

While both used perception to understand the sample contents, they still operated the robotic manipulator in an open-loop manner, without any force, visual or tactile feedback.
Lim et al.~\cite{Lim2021} also used a robotic manipulator for autonomous chemical synthesis, where they demonstrated how the robot could successfully carry out a Michael reaction with a yield of 34\%, comparable to that obtained by a junior chemist.
Nonetheless, the robotic operations were yet again simplified to pick-and-place tasks without sensory feedback.
The automation of laboratories for life sciences has perhaps a longer history of using robotics and automated platforms.
As a result of stricter protocols when compared to other fields, several works have demonstrated the usage of dual-arm robots~\cite{Fleischer2021},~\cite{Fleischer2016}, ~\cite{Joshi2019} for sample preparation and measurement.
While the robots manipulate complex tools such as pipettes, the researchers focus more on automating the instrument and hence, leave the robotic control in open-loop.
Alternatively, using multi-modal sensory feedback could potentially have not required as much modification of the instruments while allowing the robot to have increased functionality such as failure detection if something goes wrong.
Existing laboratory automation for pharmaceutical applications normally operates in using a dedicated, specific method, for example during the COVID-19 pandemic automated workflows for sample testing allowed a site to move from 180 tests per day to over 1,000 tests per day after switching to a fully automated workflow \cite{amen_blueprint_2020}. 
However, this type of automation uses specialist equipment to focus on a single task which is very different to the general use of an R\&D laboratory in which small changes will be made between batches and the workflow may rapidly change from day to day.
Another application of laboratory automation is for drug discovery where Pickles et al.~\cite{Pickles2022} demonstrate how a mobile manipulator can be used for crystallisation workflows.
The robotic platform mainly transports samples between different stations and rely on onboard sensory information such as LIDAR for navigation, while operating the robotic manipulator without visual or tactile feedback.
From all of the aforementioned works, it is evident that a vital process in laboratory automation is manipulation of vials, whether it is for sample preparation, measurement or transportation.
This is a common manipulation task across the different fields of materials discovery, life sciences, drug discovery, amongst others.
Our work addresses this task since it is fundamental across all domains and success here would scale to any workflow.
In the following sections, we will demonstrate the role multi-modal sensory feedback plays in this task, and our goal is to then transfer this knowledge to other manipulation operations in laboratory automation.

\subsection{Multi-modal Object Interaction}
Recent works also highlight the ability for a multi-modal system to significantly outperform the single modality baseline on grasping tasks, one such work shows vision guided tactile sensing can improve grasping success rate from 38.9\% to 85.2\%~\cite{jiang_where_2022}. Multi-modal sensing is also found to be a significant factor in improving a robotic cable following task~\cite{pecyna_visual-tactile_2022} where the success rate of a single modality was at maximum 77\%, however, multi-modal sensing showed a maximum of 92\%. This sensor fusion will be important in an environment where human-robot collaboration takes place as it allows the robots to form a clearer model of the environment~\cite{hansen_visuotactile-rl_2022}, better recover from failures, and avoid unwanted contact. However, this field is developing and to our best knowledge there are no works studying multi-modal feedback in the context of laboratory automation.

%% file: new/method.tex
In this work we evaluate two multi-modal methods for completing a routine laboratory vial insertion task against a single modality baseline, an overview of each modality is shown in Figure~\ref{fig:flowchart}. We divide the vial insertion task into two sub-problems:
\begin{enumerate}
    \item[i.] \textbf{Goal Position Detection:} we locate the target rack which may have been placed anywhere in the workspace. We use the circular Hough Transform to generate a large number of possible locations, a CNN then filters for vacant locations belonging to the target rack. 
    \item[ii.] \textbf{Vial Insertion:} a vial is collected from a known location and moved above a point detected by the first part of the pipeline, we then use an efficient search method and multi-modal feedback to place it into the rack.
\end{enumerate}

\subsection{Goal Position Detection}
\subsubsection{Candidate Detection}
The workspace is imaged top-down using the robot mounted camera, the circular Hough Transform (CHT)~\cite{nixon_5_2020} is applied to the image and a set of possible centre locations and radii are generated. Where the rack is in an oblique view, the CHT may fail to detect the now elliptical slots, we therefore select the detection parameters for the CHT to be overly sensitive so in these oblique cases a centre point is still detected. As consequence many image features are now detected as possible placement locations. 

\subsubsection{Filtering}
To filter the candidates we introduce a CNN classifier. Input data is created by scaling the candidate's corresponding radius by a margin factor to 110\% of the original and crop this region from the image, centred on the candidate location. The network itself is structured as an convolutional layer with a $5 \times 5$ feature map and stride of 2, followed by $2 \times 2$ max pooling, these two layers are repeated twice with independent weights and followed by a flattening layer and 3 fully connected layers with 512, 128, and 2 output neurons respectively. The network first aims to predict if the cropped image belongs to the rack and then in the case when the first parameter is sufficiently high if the rack slot is occupied. 

Training data is manually labelled from frames extracted from a video feed of the robot mounted camera moving around the workspace, labels are generated quickly via button prompt where 3 options are presented (1. Not In Rack, 2. In Rack but Occupied, 3. In Rack and Unoccupied) 

We select an insertion target from the filtered candidates by selecting the highest CNN classification score for `In Rack' and `Unoccupied', in case of a draw we select randomly from the tied candidates.

The selected target is represented by the image coordinate $(u,v)$, we then use the camera's calibrated intrinsic parameters (focal length $(f_x, f_y)$ and principal point $(c_x, c_y)$) and the height of the rack from the table $(r_z)$, and cameras position w.r.t the robot base frame $(\mathrm{CAM})$ are then used to calculate the real world position $ (x, y, z) $ w.r.t. the robot base frame:

\begin{equation}
    \begin{pmatrix} x\\y\\z \end{pmatrix} = \begin{pmatrix} \mathrm{CAM}_x + \frac{(u - c_x)(\mathrm{CAM}_z - r_z)}{f_x} \\[0.75em] \mathrm{CAM_y} + \frac{(v - c_y)(\mathrm{CAM}_z - r_z)}{f_y} \\ r_z \end{pmatrix}
\end{equation}

Finally, the vial is grasped from above (perpendicular to the workspace plane) from a second rack in a known position and lifted above the calculated target position. The multi-modal nature of the following methods relies on using this initial target position and refining it using another sensory input, either force feedback or tactile feedback. The visual single modality baseline takes a second imaging step to refine the initial estimate rather than relying on another sensor modality.

\subsection{Vial Insertion}
The second sub-task begins with the vial grasped in a two finger gripper and held above the predicted insertion location, at this point one of the three following modalities is used to insert the vial
\subsubsection{Visual Baseline}
To aid in visual only insertion we first move the camera in the $-z$ direction bringing it closer to the rack, a second image is captured and the image processing from the previous stage is repeated, the closest insertion point to the centre of the new image is selected. This process enlarges the rack in the image and ensures any skew distortion is minimal, minimising the error in the visual insertion point prediction. The vial is aligned with the revised insertion point and moved directly down the $z$ axis until it is below the rack height, at which point the gripper release command is sent and success is evaluated. 

\subsubsection{Force and Visual Feedback}
To assess the forces experienced by the vial we use the robotic arm's internal force sensors, however, the method makes minimal assumptions regarding the source of this data. We firstly initialise a first-in first-out (FIFO) buffer with length equal to one second of data from the sensor while the robotic arm is stationary, the average of this buffer then serves as a zero point to counteract the static force experienced by the sensor. This static force will vary with robot pose and payload, therefore, taking a practical measurement is preferred and does not significantly impact the speed at which the task is performed.

The robot then attempts to insert the vial by moving along the $-z$ axis, low acceleration values reduce jerk forces lowering extraneous noise in the force sensor. The FIFO buffer is monitored against the recorded static value, a deviation of more than 20\% causes the robot to stop moving and assess the vial's state. We have now reached the `Vial placed?' condition in Figure~\ref{fig:flowchart}, this is assessed using the position of the vial at the tip of the gripper w.r.t the robot base frame ($\mathrm{GRIP}$) and the height at which the vial is gripped ($V_h$). We evaluate $\mathrm{GRIP}_z >= r_z + V_h$ to indicate if the vial has impacted the top surface of the rack, creating a force which stopped the robot. In this case we move to the search algorithm detailed in Section~\ref{sec:search}, in the case the condition is not met (i.e. the vial is lower than the rack height, and thus must be in the slot) we send the gripper release command and return the robot to the starting position clear of the rack.

We select the stop condition as a 20\% deviation as the vials being handled are not excessively fragile and the robots internal force sensor can be more noisy than a dedicated sensor. This parameter can be tuned based on the tolerance for force of the glassware being handled, and the noise floor of the tactile sensor being used. The buffer length can also be varied, longer buffers exert more force on the objects as the robot will press for longer before stopping but are less prone to noise caused by jerk / joint movements, shorter buffers are more prone to false positives but will exert a smaller force on the objects involved as the system will react faster.

A second safety condition will also stop the robot in the case $\mathrm{GRIP}_z < \frac{1}{2}r_z$ to prevent the vial impacting the table should the robot misidentify a placement location and could hit the table. This condition is ultimately optional as the force feedback upon the vial applying a large force onto the table should trigger a stop, however, considering the environment and the hazards of unexpected contact this extra condition helps protect any delicate glassware from impacting an unknown object.

\subsubsection{Tactile and Visual Feedback}
For tactile feedback we rely on a pair of DIGIT~\cite{lambeta_digit_2020} visual tactile sensors, however, this method should be applicable to any pair of visual-elastomer style tactile sensors. During the initial setup phase we capture a set of reference tactile images $R$ with the gripper open and not in contact with any object, during the insertion attempt we poll the sensor at 60Hz calling each image received $i$ and applying the following processing to extract the object location. Initially we calculate the absolute difference between $i$ and each image in $R$, then average this new set of differences to produce a lower noise difference image $\delta$:
\begin{equation}
    \delta = \frac{1}{\lvert D \rvert} \sum_{d_i \in D} d_i,
\end{equation}
\begin{equation}
    \text{where}\ D = \{\ \lvert s - i \rvert,\ s \in R\ \}.
\end{equation}

The difference image is then normalised before a threshold ($0 \leq t \leq 1$) is applied, which results in a binary image $b$ where each pixel $b_{(x,y)}$:
\begin{equation}
    b_{(x,y)} = \begin{cases} 
      0 & \hat{\delta}_{(x,y)} < t \\
      1 & \hat{\delta}_{(x,y)} \geq t \\
   \end{cases},
\end{equation}
\begin{equation}
    \text{where}\ \hat{\delta} = \frac{\delta - min(\delta)}{max(\delta) - min(\delta)}.
\end{equation}

Contact regions are extracted from the binary image using the border following method detailed by Suzuki et al.~\cite{suzuki_topological_1985} which produces a set of polygons $P$ enclosing the contact regions. 
Green's theorem~\cite{duane_q_nykamp_using_nodate} is applied to calculate the area enclosed by each polygon and any excessively small regions are filtered out giving the set of contact regions $P'$. 
The centre point $(g_x, g_y)$ of each region is used to track its general position in the frame, $p$ represents a single polygon, and $v$ a vertex within the polygon $p$:
\begin{equation}
    \begin{pmatrix} g_x\\g_y \end{pmatrix} = \frac{1}{\lvert p \rvert} \begin{pmatrix} \sum v_x \\  \sum v_y \end{pmatrix}, \text{where each}\ v \in p\ \text{and}\ p \in P'
\end{equation}

Before we move the vial, its neutral position is calculated in the tactile image space, we can then use the tactile-physical mapping calculated using the method in Section~\ref{sec:mapping} to correct for a vial which has not been grasped centrally in the gripper. Initially this is generally not important as the vial is collected from a known location, however, the tactile sensors surface has a significantly lower friction than the rubber grippers. Failed placement attempts may twist the vial and the mapping gains importance as we can correct for this cumulative error.

We then move the vial down the $-z$ axis and monitor the vial's position for deviation from the neutral state. A fixed threshold triggers the robot to stop and the position is assess using the same method as the previous force feedback modality, also testing the `Vial placed?' condition in Figure~\ref{fig:flowchart}.

\begin{figure}[b]
    \centering
    \includegraphics[width=1\linewidth]{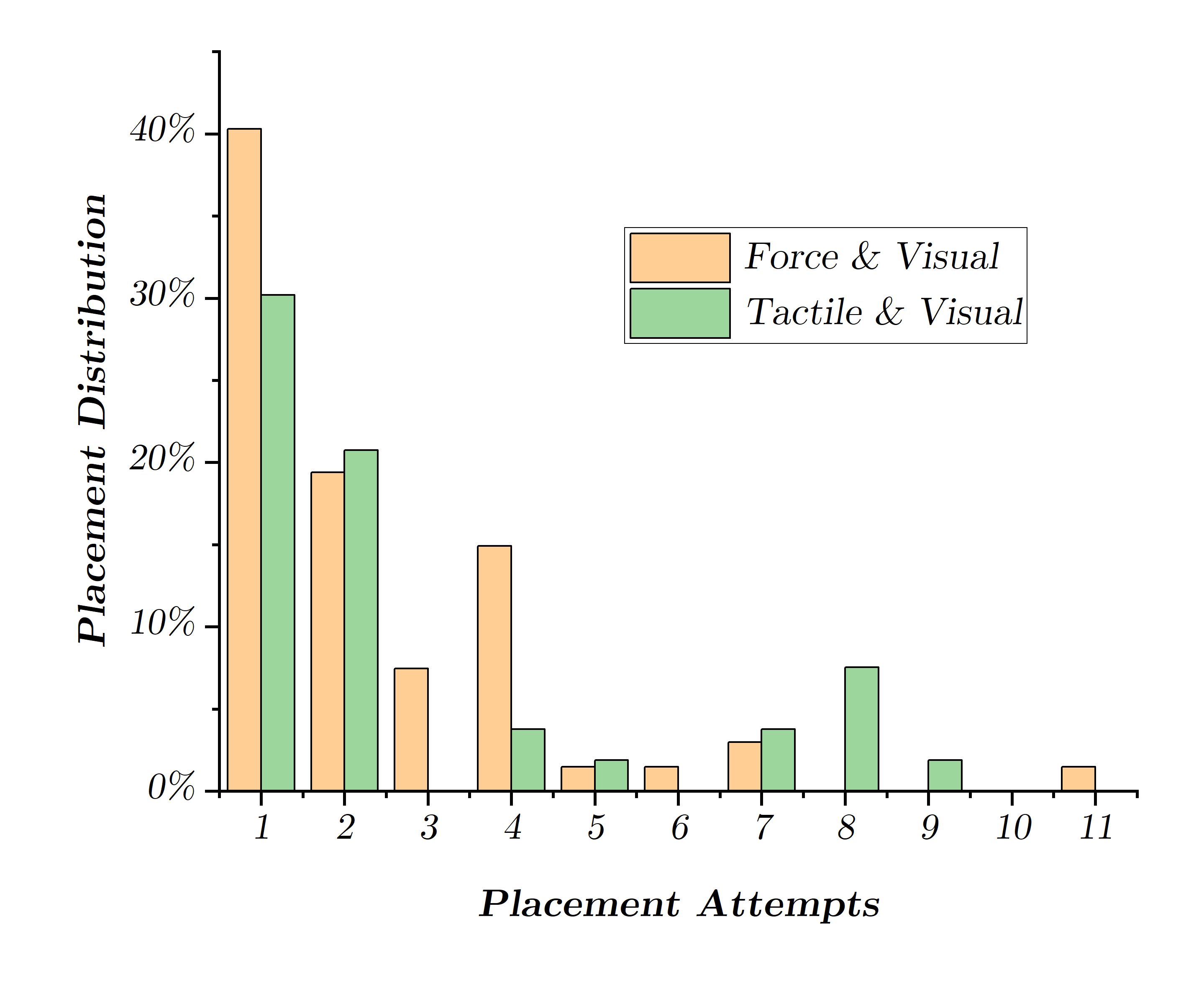}
    \caption{Distribution of successful placement attempts by number of placement attempts for the multi-modal methods}
    \label{fig:barprob}
\end{figure}

\subsection{Search Algorithm}
\label{sec:search}
The initial stage locates multiple placement positions. Using the knowledge of our selected target ($r_x, r_y$), we attempt to detect neighbours by matching a set of up to 8 other detected regions, disregarding occupancy. These neighbours provide bounds for the search we fit a bounding box centred on the selected target with the width and height denoted as ($r_w, r_h$) therefore spanning $r_x \pm \frac{1}{2}r_w$, $r_y \pm \frac{1}{2}r_h$.
It is assumed the correct placement coordinate is in close proximity to the estimate, if we deviate by more than this perimeter while searching we may accidentally insert the vial into a different slot in the rack. 
Generally, the search creates an envelope around the initial placement position at a distance represented by the search spacing. This creates possible trial locations to attempt to insert the vial into the originally targeted slot. If the vial fails to be inserted after searching the current envelope we expand it by the search spacing. Each time only the edge of the envelope is searched as its contents have been searched by previous iterations. We repeat this until either the vial is successfully inserted, or the search region is entirely beyond the bounds calculated previously.
Mathematically, we use the initial expansion factor $E = 1$, search spacing $S = 2.5\mathrm{mm}$, and occupancy set $V = \emptyset$. The expansion along the x and y axis is represented by $e_x$ and $e_y$ respectively, trials positions ($t_x, t_y$) are generated as follows:

\begin{equation}
    \begin{pmatrix}
    t_x \\ 
    t_y
    \end{pmatrix} =
    \begin{pmatrix}
    r_x \\ 
    r_y
    \end{pmatrix} +
    S \cdot \begin{pmatrix}
    e_x \\ 
    e_y
    \end{pmatrix}
\end{equation}
\begin{center}
    where $e_x, e_y \in \mathbb{Z}\ \colon e_x,e_y \in [-E,E]\ \colon \begin{pmatrix}
    e_x \\ 
    e_y
    \end{pmatrix} \notin V$
\end{center}

\subsection{Tactile Sensor Calibration}
\label{sec:mapping}
Variations between tactile sensors necessitate a practical method for calculating the transformation between a detected feature in the tactile image space and the physical position on the sensing surface. 
In our experiments we have only considered the position in the plane of the tactile sensor's contact surface, the equal pressure applied by both sides of the parallel gripper will keep any error in the gripping plane to a minimum. 
A vial is placed into a tight fitting rack at a known location, the robot arm is then positioned above the vial in the same top down grasp that will be used in later experiments. At known offsets from the vial's position the gripper is closed and a tactile image processed to find a position in the image space with a known offset on the physical sensing surface. Given varying resolutions of tactile sensors we normalise the in-image coordinate to the range $[0\dots1]$ with $0$ representing the left extreme of the sensor and $1$ the right. 
The ordinary least squares method is then applied and its parameters recorded for inference during the later experiments.

%% file: new/experiments.tex
In this section, we conduct three experiments to evaluate the three modalities defined priorly. The goal of these experiments are to investigate the performance of multi-modal sensing on the vial insertion task compared to the single modality baseline. 

\subsection{Metrics}
We consider average success rate $(\frac{successes}{attempts})$, average number of attempts before success $(\frac{\Sigma placement\ attempts}{successes})$, and the average time elapsed before success $(\frac{\Sigma time\ taken}{successes})$. An attempt is marked successful if the vial is fully inserted into the rack before being released. 

\subsection{Experiment Setup}
As shown in Figure \ref{fig:setup}, our system is comprised of a UR5 robotic arm which has a Robotiq 2F-85 gripper mounted on the end effector plate and an Intel D415 camera mounted on the `bottom' of the wrist as shown in Figure~\ref{fig:setup}. The gripper's standard rubber fingertips are used for both the visual and force feedback experiments, and they are replaced with a pair of Digit sensors~\cite{lambeta_digit_2020} for the tactile experiment.
The tactile-physical mapping and the D415 camera are individually calibrated before any experiment begins and the same calibration is used throughout all experiments, our process of calibrating the digit sensor is described in Section~\ref{sec:mapping}.
Control of the UR5 is performed via PID in both target position and target velocity modes, inverse kinematics are provided by the built-in URScript methods. 

Each experiment begins with the robot in a home pose above the workspace and a timestamp is recorded. The imaging step is then performed, and a single modality is used to insert the vial into the rack. A second timestamp is recorded when the gripper releases the vial, marking the end of that trial in the experiment. Success of the trial is evaluated and the vial is reset to the collection point, the rack is moved in the workspace and a new trial begins. 

\subsection{Results}
\begin{table}[t!]
\centering
\vspace{9pt}
\begin{tabular}{m{1.78cm} m{1cm} m{1cm} m{1.1cm} m{1.4cm}}
\toprule
Method                  & Attempts       & Runtime (s)    & Success Rate (\%) & First Time\newline Success (\%) \\ \midrule
Visual                  & $1.00\pm\newline 0.00$     & $46.41\pm\newline 1.64$ & $48.78\pm\newline 15.30$    & $48.78\pm\newline 15.30$        \\[1em]
Visual \&\newline Force Feedback & $2.4\pm\newline 1.91$ & $34.31\pm\newline 6.39$ & $89.55\pm\newline 11.56$    & $40.30\pm\newline 0.30$          \\[1em]
Visual \&\newline Tactile        & $2.86\pm\newline 2.59$ & $38.93\pm\newline 14.21$ & $69.81\pm\newline 11.48$    & $30.19\pm\newline 0.75$          \\ \bottomrule
\end{tabular}
\caption[Results for each experiment]{The number of attempts, runtime (total from initialisation to insertion), success rate and the first-time success rate for each method.}
\label{tab:results}
\vspace{-15pt}
\end{table}

\begin{figure}[b]
    \centering
    \includegraphics[width=1\linewidth]{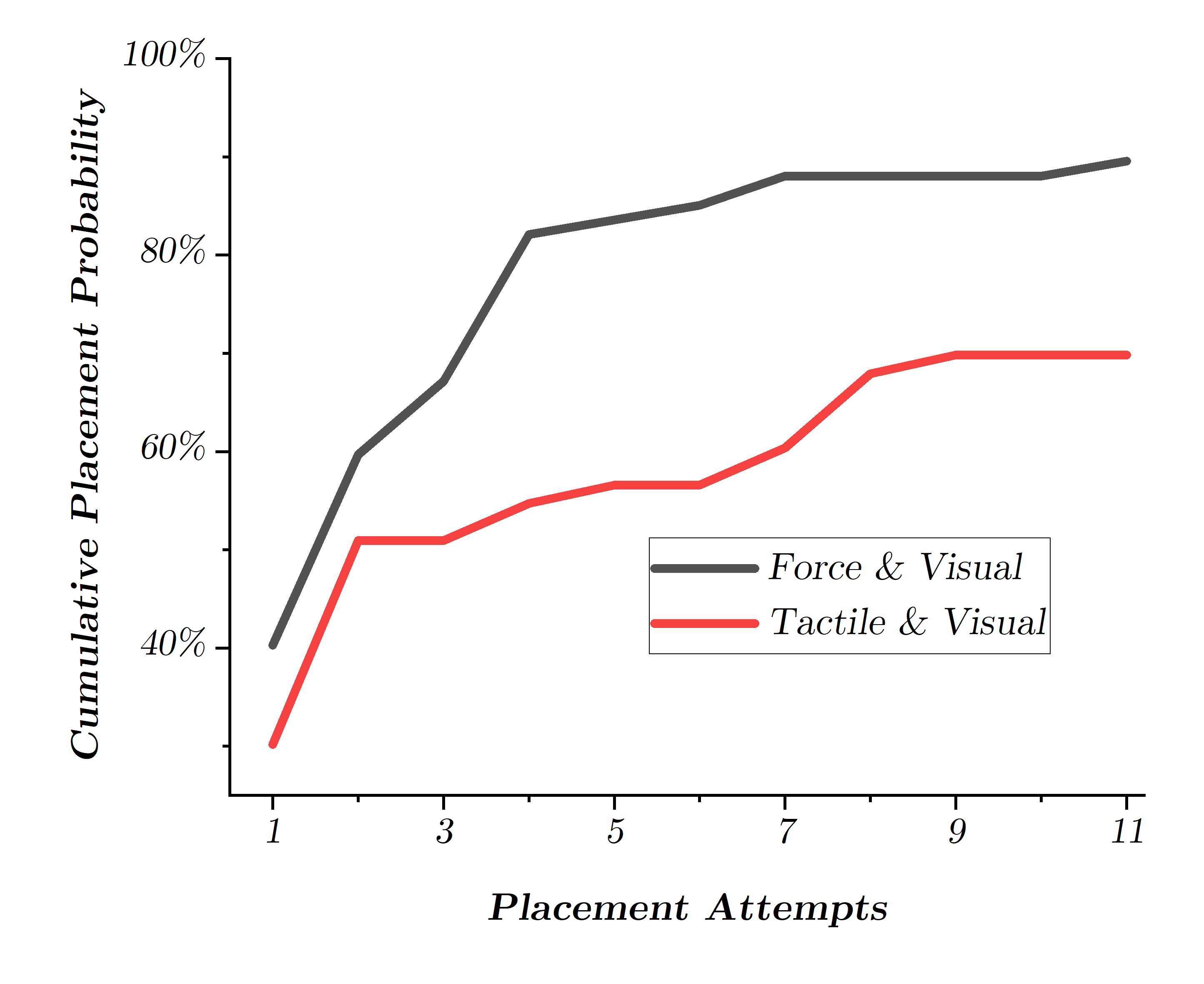}
    \caption{Cumulative probability of all attempts achieving successful vial insertion, the visual baseline sits at 48.78\% outperforming the multi-modal systems until their second placement attempt is made}
    \label{fig:lineprob}
\end{figure}

Table~\ref{tab:results} shows numerical results for each modality, Figure~\ref{fig:barprob} shows the distribution of successful placements against number of attempts, and Figure~\ref{fig:lineprob} shows the cumulative probability of the vial being placed successfully given a number of attempts.

\subsubsection{Visual Modality}
Our visual modality only makes a single placement attempt due to the limitations of our rack detection method, the transparent vial causes significant problems for visual servoing and is a difficult problem in itself~(\cite{Fang2022,Haoping2021,liu_keypose_2020,jiang_a4t_2022}). For this reason we cannot detect a failed placement attempt before releasing the vial and re-imaging the workspace. 

The baseline does not show the high reliability needed for a high throughput automation environment, with 48.78\% success. This modality also generally takes a longer time to execute, however, it's constant runtime does make it faster after the 7\textsuperscript{th} placement attempt in the alternative modalities. The target velocity of the robot arm is constant between modalities to allow this direct comparison, however, the force feedback modality does lower the allowed acceleration to this velocity to minimise jerk forces.

\subsubsection{Visual \& Force Feedback}
From Table~\ref{tab:results} the reduction in setup time by avoiding the additional visual estimate refinement step and related reduction in first time placement success is demonstrated. However, even with the additional placement attempts required the average run time is significantly lowered. This modality also displays the highest success rate of all tested modalities at 89.55\%, and as shown in Figure~\ref{fig:lineprob} it plateaus at a higher number of placement attempts, only showing minimal improvement in placement success after 4 attempts.

The decrease in first time placement accuracy shows the effectiveness of the second imaging step in the visual only modality, consider the experiment up until the first force feedback interruption, the only difference between the force feedback and visual only modalities at this point is the second imaging step. Therefore, the 8\% first time success rate lost by the force feedback method must come from the missing second imaging stage.

As this method relies on the robot's inbuilt force sensor there is potential for large forces to be generated while placing the vial, while requiring minimal modification to the robot these sensor readings also contain the static holding forces the robot is experiencing creating a significant noise floor. A larger force is then required to avoid false positives, this is undesirable in a laboratory setting as accidentally disturbing the rack could lead to a chemical spill from the sample vials contained within. Also in human-robot collaboration there is always cause for concern due to the injury risk caused by excessive force. Nevertheless, this method should be directly applicable to external force sensors which are mounted to the gripper rather than strictly using the robot's inbuilt sensing capacity; with a more accurate sensor the risks can be greatly reduced.

\subsubsection{Visual \& Tactile Feedback}
With an average runtime similar to the force feedback modality we can see that the initial reference image collection and tactile image processing introduces minimal overhead. However, surprisingly the success rate is significantly lower than the force feedback modality although still an improvement from the baseline. 
We have identified several factors which may contribute to this result: the Digit sensor cannot grasp the vial with as much force as the hard rubber gripper used in the other modalities, and its surface creates less friction due to the silicon membrane; leading to the vial moving in the gripper when a placement attempt fails, making subsequent attempts less likely to succeed even with our correction attempts.
However, this weakness also helps eliminate the safety concerns in the previous modality as the forces exerted on the vial and surfaces in contact are significantly lower. We also have more insight into the orientation of the vial, spills can be avoided as additional constraints can be introduced when planning to keep the vials orientation within tolerance.

\subsubsection{Search Method}
If the tolerance between the vial and rack is minimal the search method becomes a failure case in our experiments. Consider the case where the robot undershoots the slot on the first attempt, but due to an overly large $S$ after an envelope expansion now overshoots the slot. With this algorithm we may fail to place the vial as the search method does not dynamically adjust $S$, creating a balance between overall placement success rate and search speed. However, in our experiment using lab hardware we found the tolerance between the rack and vial to be comparatively large, aiding a human chemist to quickly handle vials without them becoming stuck in the rack, and allowing an $S$ which favours placement speed.

\subsubsection{Discussions}
The additional time per placement attempt is linear and the visual method only becomes faster after 7 placement attempts, however this is highly dependent on the acceptable speed or safety parameters for the robot. However, the visual method uses more initial robot movements therefore will be slower comparatively on the same platform compared to the other methods.
Both feedback based systems show potential for error recovery inside a robotic chemistry system. Despite undesirable force exerted the force feedback method still shows significantly improved reliability which is an important step toward closing the control loop in laboratory automation. A robotic system with multi-modality for error detection alone is a step beyond a completely open loop system which may be prone to undetected failures. However, some methods of achieving multi-modality may also inadvertently introduce new failure cases. For example, the tactile system significantly lowers the first time success rate. This may be caused by the change in material properties of the grippers, the smoother contact surface of a vision-based tactile sensor may allow the vial to slide or rotate causing a collision at the entrance to the rack. 

%% file: new/conclusions.tex
In this paper we have shown the effectiveness of multi-modal sensing for increasing reliability in a common laboratory vial handling task. We have also introduced an effective filtering method for identifying valid placement locations by combining existing classic computer vision methods and machine learning, and a bounded search method for recovering from a failed placement attempt. Both our multi-modal approaches show significant improvement over the single modality approach and the increased richness of the sensory data offers additional attractive features in the laboratory automation context. 
Despite offering a smaller improvement over the baseline inclusion of tactile sensors allows for improved safety in the movement planning stage via additional insights into the grasped objects orientation. However, the inclusion of a standalone force sensor also provides greater insight about the external contact forces being placed on the grasped object as the tactile approach generally requires slippage to occur before force is detected using our method.
Albeit showing the best placement reliability the force feedback system without a standalone sensor still presents some concerning properties for the human-robot collaborative environment, such as excessive force being applied before a stop is triggered.
A possible improvement to the search algorithm would explore from the maximum bounds inward, dynamically modifying the search spacing similar to a binary search. Avoiding an excessively large $S$ causing failure cases. Accounting for the previous search results may also allow the search to be biased towards an area, over time offsetting any calibration error between the gripper and camera or tactile sensors.
In future works the combination of all three modalities could be explored for a combination of the greater placement ability of the force feedback system and also the increased insight into the grasped objects pose from the tactile system. We also believe the use of machine learning could further increase the placement success rate by employing a reinforcement learning approach potentially using all 3 modalities simultaneously in a sim2real approach.